\documentclass[conference]{IEEEtran}
\IEEEoverridecommandlockouts
\usepackage{cite}
\usepackage{amsmath,amssymb,amsfonts}
\usepackage{algorithmic}
\usepackage{graphicx}
\usepackage{textcomp}
\usepackage{xcolor}
\usepackage{hyperref}
\usepackage{setspace}
\usepackage{etoolbox}
\usepackage{listings, lstautogobble}
\usepackage{url}

\AtBeginEnvironment{quote}{\singlespacing\small}
\AtEndEnvironment{quote}{\bigskip}

\def\BibTeX{{\rm B\kern-.05em{\sc i\kern-.025em b}\kern-.08em
		T\kern-.1667em\lower.7ex\hbox{E}\kern-.125emX}}

\begin{document}
	
	\title{Applied Machine Learning Methods with Long-Short Term Memory Based Recurrent Neural Networks for Multivariate Temperature Prediction\\
	}

	\author{\IEEEauthorblockN{Bojan Lukić}
	\IEEEauthorblockA{\textit{Clausthal University of Technology}\\
	\textit{Institute for Software and Systems Engineering}}\\
	}

	\maketitle
	
	\begin{abstract}
	This paper gives an overview on how to develop a dense and deep neural network for making a time series prediction. First, the history and cornerstones in Artificial Intelligence and Machine Learning will be presented. After a short introduction to the theory of Artificial Intelligence and Machine Learning, the paper will go deeper into the techniques for conducting a time series prediction with different models of neural networks. For this project, Python's development environment Jupyter, extended with the TensorFlow package and deep-learning application Keras is used. The system setup and project framework are explained in more detail before discussing the time series prediction. The main part shows an applied example of time series prediction with weather data. For this work, a deep recurrent neural network with Long Short-Term Memory cells is used to conduct the time series prediction. The results and evaluation of the work show that a weather prediction with deep neural networks can be successful for a short time period. However, there are some drawbacks and limitations with time series prediction, which will be discussed towards the end of the paper.
	\end{abstract}
	\bigskip
	
	\section{Introduction}
	When talking about Artificial Intelligence (AI) and Machine Learning (ML), many people think of a technology that has only emerged recently and is built on futuristic machines or robots. However, the idea of Machine Learning and neural networks in the context of Artificial Intelligene has already emerged in the 19th century. In 1943, Warren McCulloch and Walter Pitts published a paper [1] with the title "A Logical Calculus of Ideas Immanent in Nervous Activity". In their work the neurophysiologist McCulloch and the mathematician Pitts present a simple model of Artificial Neural Networks (ANN) for complex computation, shown on the example of animal brains. This landmark model for neural networks was the first of its kind and many similar ones followed after it.
	
	ML, a subset of AI, has been used in practical applications since the 1990s. A very popular example of an ML application is the spam filter. Of course, a spam filter might not seem as intelligent as today's AI applications such as Google's Deep Mind. Nevertheless it qualifies as ML, considering how well spam filters can filter harmful emails today. The success of ML with spam filters has led to a whole slew of new applications using ML to power products [2].
	
	One of these applications that poses another disruptive technology, is the deep neural network developed by Geoffrey Hinton et al. Their paper [8] discusses a method of building a dense belief net with multiple layers for recognizing handwritten digits with an (at that time) unprecedented precision of more than 98\%. Most scientists had dropped the idea of deep networks around the turn of the millenium. This groundbreaking paper reawakened the interest for research in the field of deep neural nets. Following papers showed success in achieving high performances with ML. 
	
	In the previous decade, ML and AI became well reputed technologies used in a lot of software and also physical products, reaching from Google's Assistant speech recognition to beating world champion's in video games such as Dota 2. Considering the fast rate of development in the sector of AI, ideas such as self-driving cars are not that far from reality anymore [2].
	\bigskip
	
	\section{Related Work}
	In the last decade, several efforts have been made to achieve precise time series prediction with weather data including ML models.
	
	In 2007, Mohsen Hayati and Zahra Mohebi presented a paper [3] in which they predicted dry air temperature. They used a structure of Perceptrons consisting of three layers using a mix of the sigmoid activation function and a linear activation function. In total, seven measures gathered over a period of 10 years were used to train and test the model: solar radiation, sunbeams, air pressure, humidity, wind speed, wet temperature, and dry temperature. After normalization, the measures were used to make 8 temperature predictions per day in even intervals. The accuracy of Hayatis's and Mohebi's model was measured in mean absolute error with an acceptable average of 1.3.
	
	Similarly, Mohamed Akram Zaytar and Chaker El Amrani presented a deep neural network architecture to make a time series weather prediction in 2016 [4]. The data consisted of the three measures wind speed, relative humidity, and temperature which was gathered over a period of 15 years. Their neural network consisted of multi stacked Long Short-Term Memory cells with the rectifier activation function and multiple hidden layers. Here, predictions were made for 24 and 72 hours into the future. The accuracy of Zaytar's and Amrani's model was measured in mean squared error. The results showed that a deep Long Short-Term Memory network is capable of forecasting general weather variables with a good accuracy.
	
	More recent papers [5], [6], [7] from 2018 and 2019 show similar approaches to time series predictions with multilayer Perceptrons and Long Short-Term Memory cells. All of the papers show very satisfying time series predictions and a potential for increasing the accuracy of the results.
	
	By now there is a vast number of ML tools available, with Python and the extension TensorFlow being a very popular pick due to their ease of use. 
	The related papers presented above showed that complex neural networks possess the capability of making satisfactory time series predictions. Especially when tapping the full potential of ML models by providing a high amount of data and tuning the parameters of a neural network, results can be above average. However, there are some limitations to ML models. One of the main limitations that this paper will try to address is the decreasing accuracy with an increase in prediction points in structures with Long Short-Term Memory cells.

	This paper's approach that addresses these existing limitations is to create a default model for temperature predictions, progressively improve the model, and find the threshold for prediction points with an acceptable accuracy.
	\bigskip
	
	\section{Background of Machine Learning and Artificial Intelligence}
	This section gives a more detailed overview on the developments and theory of Machine Learning and Artificial Intelligence with neural networks.
	\bigskip
	
	\subsection{Machine Learning}
	So what exactly is ML? A quote from Tom Mitchell's publication [9] on ML summarizes the term quite well:
	
		\begin{quote} 
		A computer program is said to learn from experience E with respect to some task T and some performance measure P, if its performance on T, as measured by P, improves with experience E. —Tom M. Mitchell, 1997 
		\end{quote}
	
	To make this definition easier to understand, one can think of an example: A computer program could improve its performance with experience gathered through playing many games against another player or even itself. The performance could be measured with the improvements the program makes over time and the number of games it wins and loses.
	
	As mentioned before, the spam filter for email services is an ML program. The way spam filters learn to filter spam from ham (ham emails are the opposite of spam emails) is by getting input from users (for example in form of flagged emails) and sets of regular, nonspam emails. The set of regular emails that the software uses to learn is called training set, which can be one of many training instances. If the definition from Tom Mitchell is used on the example of spam filters, E would be the training set, T the task of branding spam mails as spam, and P a performance measure such as the success rate of flagging spam mail. 
	
	Looking at figure 1, in the Machine Learning approach, a spam filter automatically gets fed training data which it uses for distinguishing between spam and ham mails. After evaluating the results of the training, the filter can either be launched or further trained, until it reaches the desired accuracy. Due to its nature of automated learning, the system is usually easier to maintain, more independent and also more accurate than a conventional spam filter.
	Also, if new types of spam emails with a new subject phrasing or content occur, spam filters using ML will automatically detect the unusually high amount of flagged emails containing these new terms. In contrast, a conventional spam filter program would need to be updated manually for new spam mails.
	
	\begin{figure}[!htbp]
		\centerline{\includegraphics[width=\linewidth]{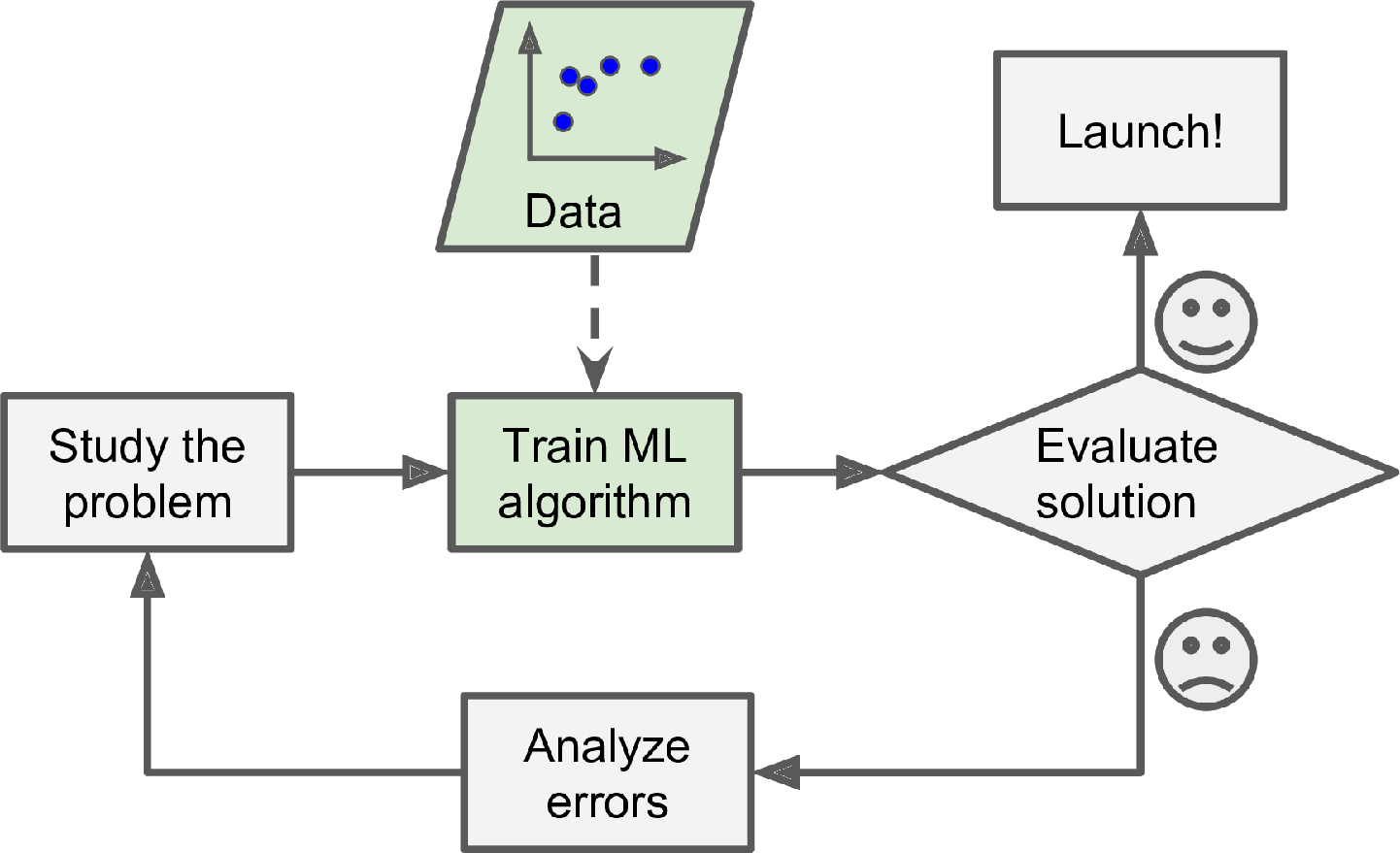}}
		\caption{The Machine Learning approach [2].}
		\label{fig}
	\end{figure}

	A counterexample of something that would not specify as ML is transfering a big amount of data to a system without further using it. The system contains much more data than before, but has not used the data for any learning tasks. Simply providing information without processing, analyzing and using it is not considered ML [2].
	\bigskip
	
	\subsection{Artificial Intelligence and Neural Networks}
	After the early introduction to Artificial Neural Networks (ANN) by McCulloch and Pitts, many researchers believed in a rapid development of AI. After two decades of slow progress and the realization that the hopes of independent and intelligent machines would not come true, other systems were developed and the work with ANNs came to rest. After two more decades of slow developments in the field of AI, other systems such as Support Vector Machines (also a type of ML), were funded and increasingly used. These alternatives offered better accuracy and a better framework than ANNs, which further slowed the progress of AI.
	Back in the 1940s, McCulloch and Pitts presented a simple, artificial reproduction of a natural neuron: the artificial neuron. An artificial neuron has at least one binary input as well as one output. In their work, it was shown that after a certain number of input activations, the neuron can fire to make an output. In figure 2, very basic ANNs with various logical prepositions can be seen. In these examples, a neuron fires after at least two input activations. McCulloch and Pitts showed that even with such a simple model, it is possible to construct an ANN which can process different prepositions.
	
	\begin{figure}[!htbp]
		\centerline{\includegraphics[width=\linewidth]{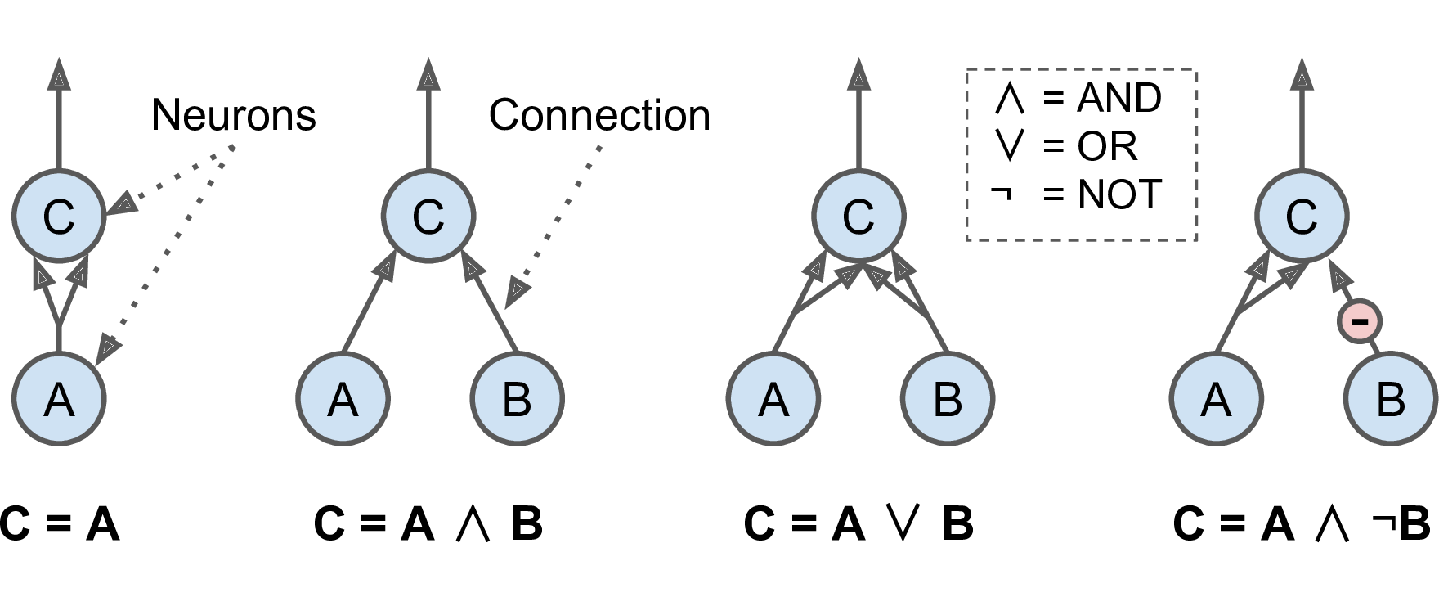}}
		\caption{ANNs performing simple logical computations [2].}
		\label{fig}
	\end{figure}

	In the first network from the left, the neuron C will only fire if neuron A fires first. Because neuron C receives two input activations after A fires, the above condition is fulfilled.
	In the second network, C will only fire if both A and B fire first. 
	The third network contains an OR logic. It is sufficient if either A or B fires. In this case, each input neuron will provide two input activations upon firing, which is sufficient for C to activate and make an output.
	The last example on the right introduces the NOT logic: C will only make an output if A fires and B does not fire. 

	A slightly different artificial neuron is the Perceptron. Frank Rosenblatt invented this very simple ANN architecture. Figure 3 shows the variation of the simple artificial neuron shown above. This model goes by the name linear threshold unit (LTU) or (more commonly) threshold logic unit (TLU). Contrary to the binary output values in the neural networks above, the outputs as well as the inputs of this model are numbers. Also, every input carries a weight. A TLU sums its weighted inputs (z = w\textsubscript{1} x\textsubscript{1} + w\textsubscript{2} x\textsubscript{2} + ... + w\textsubscript{n} x\textsubscript{n} = x\textsuperscript{T} w), before applying a function and outputting a result: h\textsubscript{w}(x) = step(z), where z = x\textsuperscript{T} w.
	
	\begin{figure}[!htbp]
		\centerline{\includegraphics[width=\linewidth]{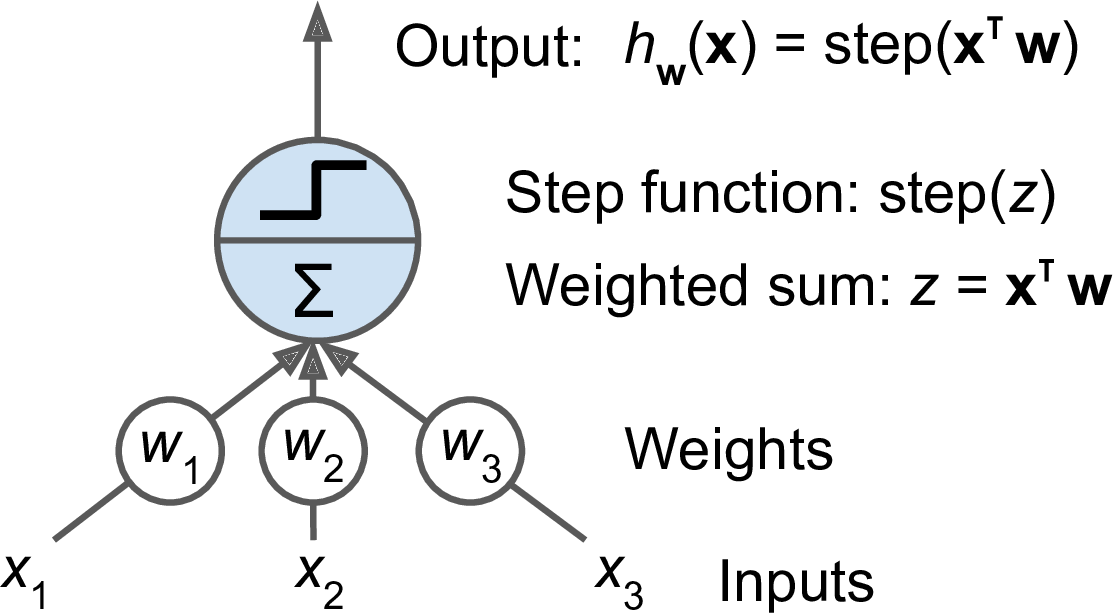}}
		\caption{Threshold logic unit: an artificial neuron which computes a weighted sum of its inputs then applies a step function [2].}
		\label{fig}
	\end{figure}

	A Perceptron contains a single row (or layer) of TLUs. Every TLU has a conncetion to all the inputs. When Perceptrons are stacked on top of each other, the neurons from one layer can be connected with the neurons from the neighboring layers. Dense or fully connected layers can be observed, when all artificial neurons in one layer are connected to the neurons from the surrounding layers. All neurons providing input activation constitute the input layer. Additionally, a bias feature is added to the Perceptron, called bias neuron. This special neuron always outputs 1. 
	The weighted sums and step function define an equation which can be used to very precisely calculate the outputs of a layer of neurons for several instances:
	
	\begin{equation*}
		h\textsubscript{W, b}(x) = \phi(XW + b)\\
	\end{equation*}
	
	In the equation, X stands for the input features. It consists of a matrix with each column representing a feature and each row showing one instance. W stands for all weights from each input into a TLU (see figure 3). The bias unit b is added seperately and its weight is not contained in W. In the weight matrix W, each column represents one neuron with one row per input neuron. Lastly, the bias unit b represents all the weights between the connections of the artificial neurons and the bias neurons. There is always one bias unit per neuron. The activation function $\phi$ can be seen as a binary. In its simplest form, $\phi$ defines if a neuron is firing or not. In this example, all neurons are TLUs which makes $\phi$ a step function (0 for negative arguments and 1 for positive arguments).
	
	In the early days of ANNs, researchers found that Perceptrons showed major weaknesses in solving even small problems. One of these problems is the exclusive or (XOR) classification problem. The XOR is a logical operation that outputs the binary true only when the system's inputs differ. This could be written as "A or B, but not, A and B". 
	As mentioned before, Rosenblatt showed in his work that a simple Perceptron is capable of giving correct outputs for simple logical operations, such as AND, OR, and NOT. However, those Perceptrons were not capable of solving the XOR classification. At that time, all of the linear classification models had difficulties with these types of classification problems, but regardless, scientists had much higher expectations of Perceptrons. This limitation to ANNs led to scientists dropping research on it altogether.
	One way to overcome these problems, however, is to stack multiple Perceptrons on top of each other. This modified ANN is called multilayer Perceptron (MLP). With an MLP, the XOR classification problem can be solved. With the marked connections in figure 4 having a weight other than 1 (-1.5, -0.5, -0.5, and -1), the network will output 0 for the inputs (0, 0) and (1, 1) and 1 for the inputs (1, 0) and (0, 1) [2].
	
	\begin{figure}[!htbp]
		\centerline{\includegraphics[width=\linewidth]{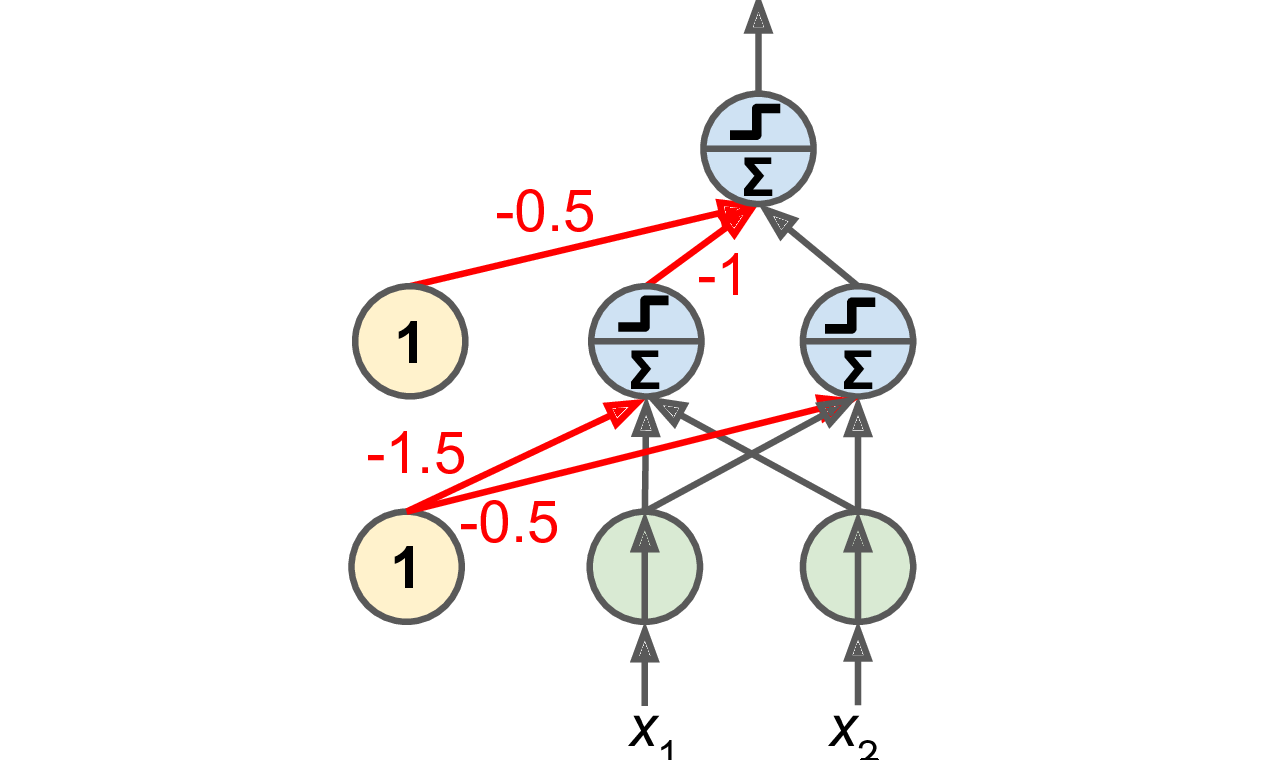}}
		\caption{An MLP that solves the XOR classification problem [2].}
		\label{fig}
	\end{figure}
	
	The success of MLPs solving more complex classification problems has resulted in the resumption of research in the field of ANNs.
	More complex multilayered ANNs, which can also be used for time series prediction will be discussed in more detail in the main part of this paper.
	\bigskip
	
	\section{Problem statement and system setup}	
	The aim of this project is to make a weather prediction, based on historical weather data gathered at the Institute of Electrical Information Technology of the Clausthal University of Technology. More precisely, the aim is to forecast the temperature for a time period of 7 days. A *.csv file contains weather data from the time period of November 1st 2003 until December 3rd 2012. Following measures are captured in the file: temperature, humidity, air pressure, solar radiation, wind velocity, and wind direction. The measures are stored as median values for intervals of one hour. This means that for one day there are 24 entrys per measure with each entry being the median value for the time between the previous and the following measurement. 
	Hereinafter, the measures are used with the following abbreviations:
	\bigskip
	
	\begin{tabular}{l}
	1. temp - median temperature per hour in this row \\
	2. hum - median humidity per hour in this row \\
	3. airpr - median air pressure per hour in this row \\
	4. solrad - median solar radiation per hour in this row  \\
	5. windvel - median wind velocity per hour in this row \\
	6. winddir - median wind direction per hour in this row \\
	\end{tabular}
	\bigskip
	
	For this time series prediction, a framework of different tools is used. The fundamental development environment is the Jupyter notebook running Python code. A virtual environment stores all the necessary packages and libraries for conducting a time series prediction. In this case the library TensorFlow for distributed numerical computation is used. TensorFlow is a very powerful library which can run complex neural networks. 
	Here, TensorFlow will be used together with Keras. Keras is a neural network library, providing an easy way to train and run neural networks. TensorFlow has its own Keras implementation, which supports TensorFlow as a backend. 
	For preprocessing the data and running specific metrics, Scikit-Learn, which features various classification, regression, and clustering algorithms will be used.
	
	The following code segment shows an example of an MLP programmed with Python in the above mentioned neural library framework:
	
	\begin{lstlisting}[basicstyle=\scriptsize,]
	import tensorflow as tf
	from tensorflow import keras
	
	model = keras.models.Sequential()
	model.add(keras.layers.Flatten(input_shape=[28, 28]))
	model.add(keras.layers.Dense(300, activation="relu"))
	model.add(keras.layers.Dense(100, activation="relu"))
	model.add(keras.layers.Dense(10, activation="softmax"))
	\end{lstlisting}
	
	In the code segment, the input layer is preprocessed by converting it into a one dimensional array with "Flatten". Two dense, hidden layers are added with the first one having 300 and the second one 100 neurons. Each layer handles its own weight matrix. The "ReLU" activation function makes sure that the outputs in the output layer will always be positive. Finally, the output layer contains 10 neurons with a "Softmax" activation function.
	
	When building an ANN, any number of layers with as many neurons as desired can be added to a model. For the time series prediction, a similar ANN structure as in the code segment above, with a slightly different type of network, will be used.
	\bigskip
	
	\section{Different approaches to time series prediction}
	To conduct an accurate and relevant time series prediction such as a weather forecast, a sequence of multiple measures is needed. Since there are multiple measures per time step, the series is a multivariate times series. Forecasting on a multivariate time series can be conducted with different methods. For this example, a variation of a Recurrent Neural Network (RNN) cell will be used, called Long Short-Term Memory (LSTM) cell.
	Before conducting the time series prediction, RNNs and LSTMs will be discussed briefly in the next subsections.
	\bigskip
	
	\subsection{Recurrent Neurons and Layers with memory cells}
	The ANNs discussed above are feedforward neural networks, meaning that the activations flow in one direction: From the input to the output layer. RNNs are similar to ANNs, however, they also have connections flowing backwards [10].
	This gives RNNs the ability to store memory while processing elements. Also, standard ANNs can only process one sequence element at a time, while RNNs have the ability to process multiple values, such as a sequence of data, more efficiently [11]. 
	The RNN's memory, called recurrent hidden state, gives it the ability to make predictions about future inputs in a sequence of input values. In theory, RNNs can therefore use sequential data to predict an indefinitely long sequence of future data. Further into this subsection it will be shown that in practice an indefinitely long prediction is not always possible [12].
	RNNs are used to process speech and text, as well as make time series predictions [10].	
	
	Figure 5 shows an example of a simple recurrent neuron (left) and an RNN (right). The recurrent neuron receives an input x, which produces an output y. The output is also sent back to the neuron itself. In addition to the input x(t) per time frame t, the neuron receives its own output y(t-1) from the previous time frame t-1. Adding an axis for time, a network can be created, which illustrates all inputs and outputs for the same recurrent neuron per time frame [2].
	
	\begin{figure}[!htbp]
		\centerline{\includegraphics[width=\linewidth]{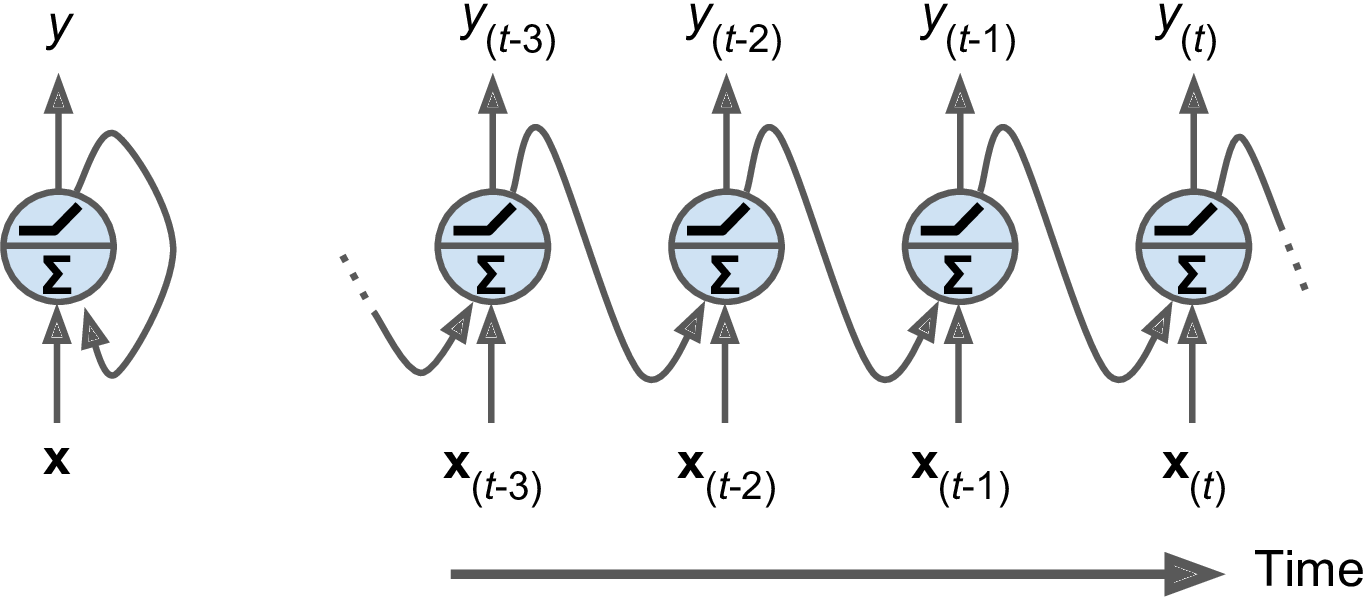}}
		\caption{A recurrent neuron (left) unrolled through time (right) [2].}
		\label{fig}
	\end{figure}

	It is possible to create layers of recurrent neurons. Similar to MLPs, recurrent neurons stored in one layer each receive the inputs per time frame x(t) as well as the outputs from the previous time frame y(t-1). Every recurrent neuron has a weight for the inputs x\textsubscript{(t)} as well as for the outputs from the previous time frame y\textsubscript{(t–1)}. As the inputs and outputs of layered RNNs are vectors, the weights for the inputs and outputs per recurrent neuron can be defined as vectors w\textsubscript{x} and w\textsubscript{y}. The weights for the recurrent neurons from one layer can be summed up into the weight matrices W\textsubscript{x} and W\textsubscript{y}. The bias unit b is summed seperately.
	The following equation shows the calculation of the output vector for one whole layer of recurrent neurons:

	\begin{equation*}
		y\textsubscript{t} = \phi(W\textsubscript{x}\textsuperscript{T}x\textsubscript{(t)} + W\textsubscript{y}\textsuperscript{T}y\textsubscript{(t-1)} + b)\\
	\end{equation*}
	
	As mentioned above, a recurrent neuron has a form of memory. A part of a neural network, which contains recurrent neurons that preserve states across time frames, is called a (memory) cell. Networks with memory capability are especially useful for time series predictions, such as weather forecasts. When the network receives historical weather data in a sequence it will output a future sequence of values [2].
	
	Unfortunately, RNNs also have some drawbacks. Common problems are the "blow up" of signals or the "vanishing" of information over a longer time period. The blow up of signals happens when the inputs in an RNN accumulate through its layers and create a very big gradient towards the end or beginning layer. This tends to build up so much that in the end, the signals blow up and make outputs useless [12], [13].
	On the contrary, "vanishing gradients" happens, when inputs travel through a great amount of layers and fade towards the end or beginning layer [12], [14].	
	\bigskip
	
	\subsection{Deep Recurrent Neural Networks}	
	A variation of the RNN is the deep RNN. As shown in figure 6, a deep RNN is simply a batch consisting of multiple layers of cells.
	
	\begin{figure}[!htbp]
		\centerline{\includegraphics[width=\linewidth]{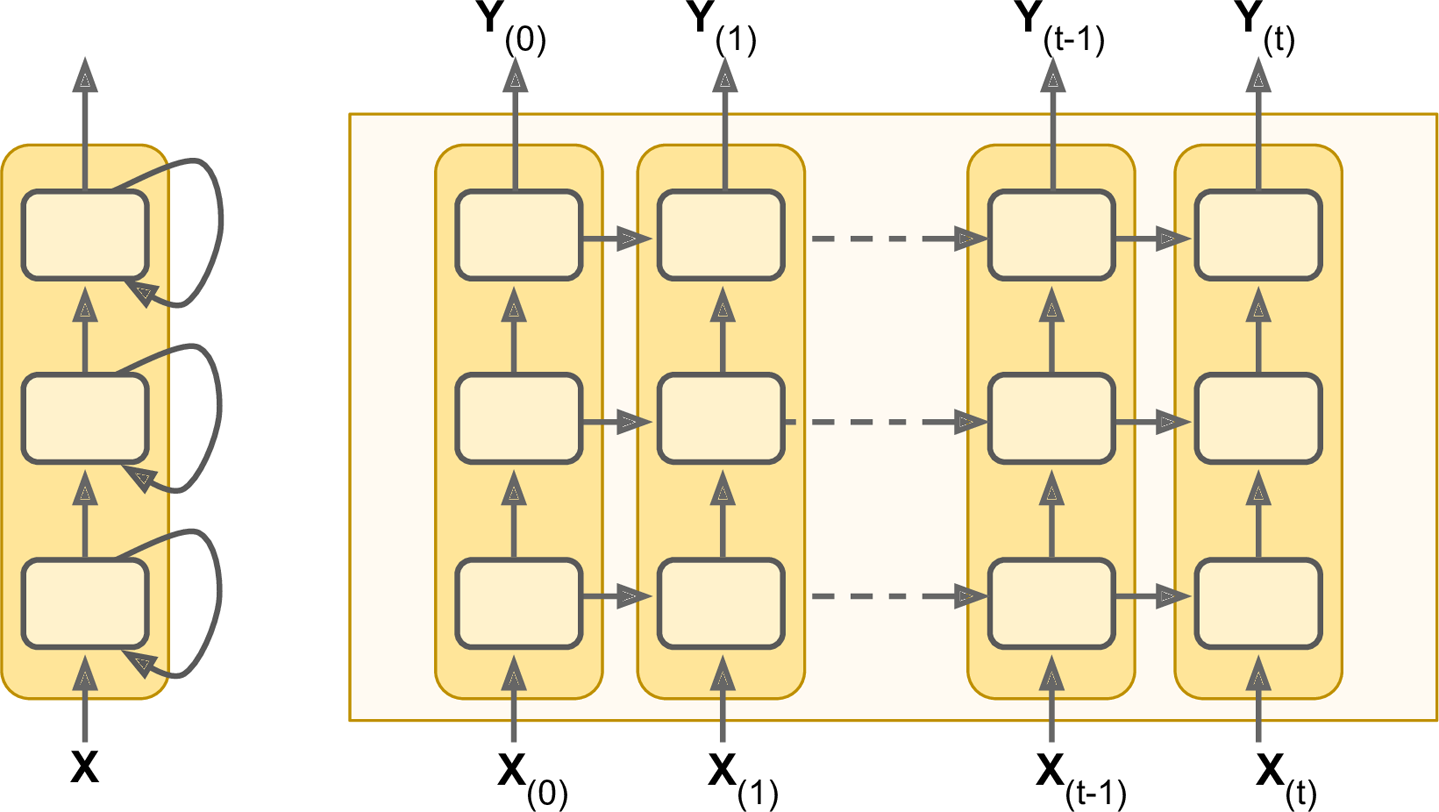}}
		\caption{Deep RNN (left) unrolled through time (right) [2].}
		\label{fig}
	\end{figure}
	
	Programming a deep RNN with TensorFlow and Keras is similar to programming an MLP, as seen in the code segment below figure 6.
	
	\begin{lstlisting}[basicstyle=\scriptsize,]
	model = keras.models.Sequential([
	keras.layers.SimpleRNN(20, return_sequences=True, 
			input_shape=[None, 1]),
	keras.layers.SimpleRNN(20, return_sequences=True),
	keras.layers.SimpleRNN(1)
	])
	\end{lstlisting}
	
	It has been shown [2], [8], [15] that with each new layer in a deep RNN, the overall model improves. Again, it is possible to stack more layers with more recurrent neurons in above example.
	
	Due to the previously mentioned transformation of data running through a (deep) RNN, information can be distorted or lost over time. In the worst case, the outputs towards the beginning or the end of an RNN do not contain any of the information provided with the first inputs. 
	Over time, new cells with a long-term memory have been developed, which have shown success in tackling the issues of traditional RNNs. One of the most popular discoveries is the LSTM cell [2]. 	
	\bigskip
	
	\subsection{Long Short-Term Memory}	
	Just like RNN cells, LSTM cells can store dependencies. Additionally, based on the information flowing through the neural network, LSTM cells have the ability of dropping information that is not useful. They will also model the relationships between the long input and output data [12]. Ultimately, the performance of LSTM cells will be better and the processing of training data will be faster than with conventional RNN cells [2].
	The loss of useless information happens in a "forget gate" of an LSTM cell. A special activation layer, called sigmoid activation layer, outputs a value of either 0 or 1. If the activation layer outputs 0, the gate forgets everything in the corresponding cell, for the output 1, all information is kept [10], [16]. 
	
	The architecture of an LSTM cell is shown in figure 7. 	
	The cell has two states h\textsubscript{(t)} and c\textsubscript{(t)}. h\textsubscript{(t)} is the short-term state and c\textsubscript{(t)} the long-term state of the cell [2].
	
	\begin{figure}[!htbp]
		\centerline{\includegraphics[width=\linewidth]{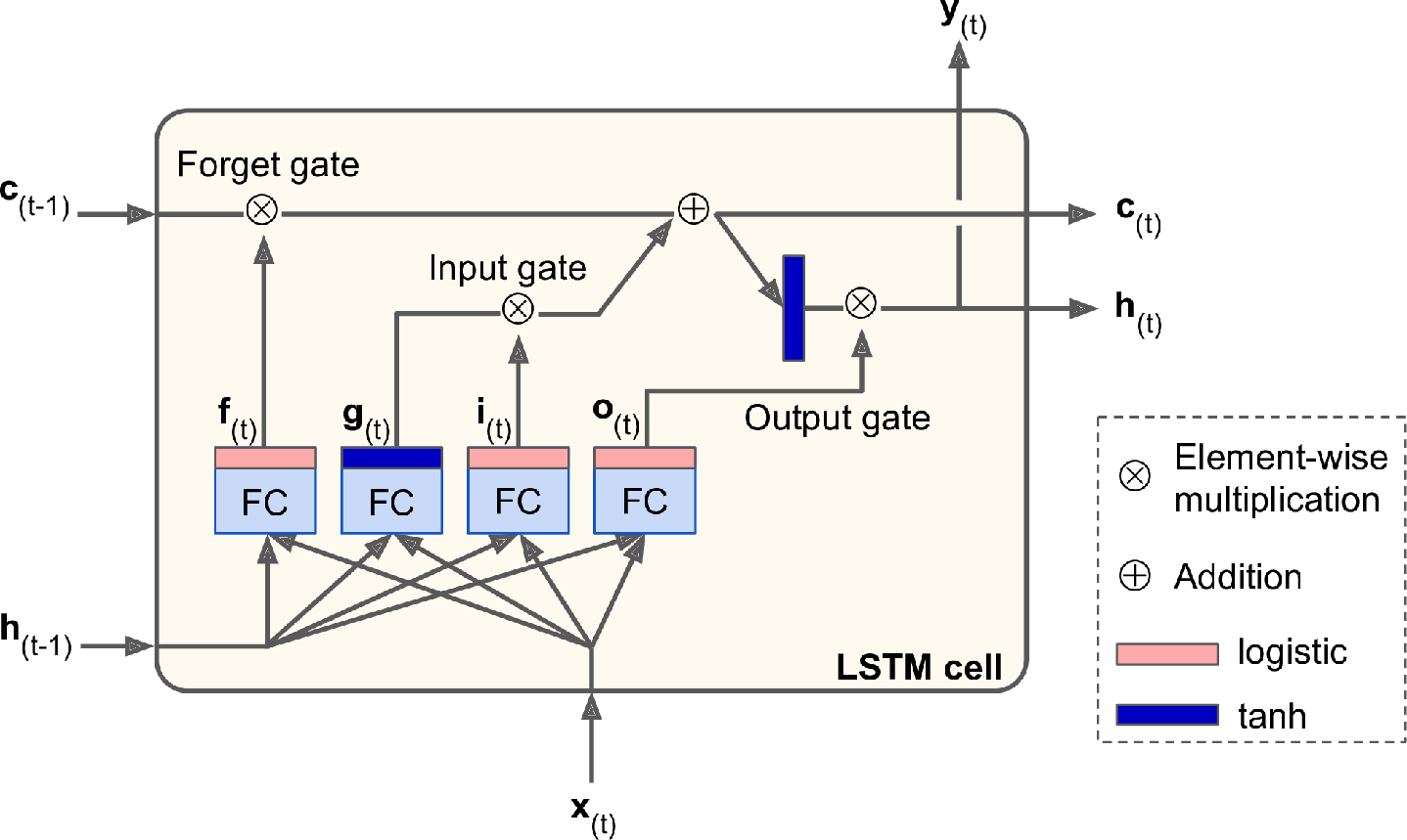}}
		\caption{LSTM cell [2].}
		\label{fig}
	\end{figure}

	The main layer in this architecture outputs g\textsubscript{(t)}. In a conventional cell, there is only the main layer which receives inputs from x\textsubscript{(t)} and the previous state h\textsubscript{(t-1)}. It then outputs information to y\textsubscript{(t)} and h\textsubscript{(t)}. 
	In an LSTM cell, the most important information of g\textsubscript{(t)} is stored in a long-term state, while information that is not useful is dropped. 
	The three other layers f\textsubscript{(t)}, i\textsubscript{(t)}, and o\textsubscript{(t)} use the above mentioned activation function, which outputs 0 or 1. The outputs of the three layers f\textsubscript{(t)}, i\textsubscript{(t)}, and o\textsubscript{(t)} cross the forget gate (for output f\textsubscript{(t)}), the input gate (for output i\textsubscript{(t)}), and the output gate (for output o\textsubscript{(t)}). All inputs that flow through the cell are multiplied with the outputs of the activation functions. 
	For example, the input of the long-term state c\textsubscript{(t–1)} flows through the network and crosses the forget gate. If f\textsubscript{(t)} outputs 0, the forget gate will be closed and memories will be dropped. If it outputs 1, the gate is opened and the long-term state keeps its memories. Via the addition operation, c\textsubscript{(t–1)} can add memories from the input gate, before being output as c\textsubscript{(t)} [2].
	
	In Keras, the SimpleRNN layer used in the previous code segment can simply be changed with an LSTM layer, creating a neural network with LSTM cells.	
	
	\begin{lstlisting}[basicstyle=\scriptsize,]
	model = keras.models.Sequential([
	keras.layers.LSTM(20, return_sequences=True, 
			input_shape=[None, 1]),
	keras.layers.LSTM(20, return_sequences=True),
	keras.layers.Dense(10))
	])
	\end{lstlisting}
	
	Due to the LSTM cells' ability of dropping memories and storing useful information over a long time period, these cells can overcome the problems mentioned with RNNs. Even when there is unintelligible sequences flowing into the neural network or noise in the information, LSTM cells can bridge time intervals efficiently and are able to tackle the vanishing gradients problem [13]. With the extension of the RNNs' memory, LSTM cells can learn and store long-term dependencies better [12], [14]. LSTM cells have shown great success in speech recognition, handwriting recognition, and time series forecasting [11]. 
	\bigskip
	
	\section{Conducting a time series prediction with the LSTM model}
	For this project, the LSTM model will be used. The capabilities the model has in detecting long-term dependencies in the data and filtering the most important inputs makes LSTM cells especially suitable for time series predictions.
	\bigskip
	
	\subsection{Setup}
	In case the data used for time series prediction is in raw form, it first needs to be prepared. The dates need to be put in the right format, such as "day, month, year, hour". Further, entries with the values "N/A" need to be replaced with "0". Any empty rows should be removed or consolidated with filled rows. Once the data has been prepared the data processing can continue.
	
	In this case, the weather data stored in the *.csv file has already been normalized and cleaned.
	The supervised learning problem can be framed with predicting the weather at the current hour (t) given the weather measurement and weather conditions at the prior time steps.
	After normalizing and transforming the data, the first 5 rows of the result can be printed: 
	
	\begin{figure}[!htbp]
		\centerline{\includegraphics[width=\linewidth]{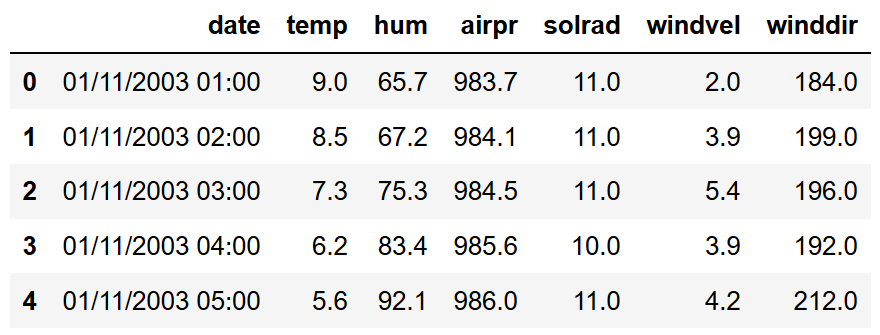}}
		\caption{First 5 rows of the data set.}
		\label{fig}
	\end{figure}
	
	It is common to use 80\% of the data for training and to hold out 20\% for testing. However, this depends on the size of the data set: if the data set contains 10 million instances, then holding out 1\% means the test set will contain 100,000 instances. That’s enough to get a good estimate of the generalization error [2].
	The model in this demonstration will be fit on the first 7 years of data, then evaluated on the remaining 2 years of data. This is a ratio of roughly 0.78 to 0.22 for training to testing data. Ideally, a temperature forecast for 7 days should be created with the model. This type of time series prediction is called multi step time series prediction, as multiple steps in the future should be predicted. The model is also multivariate, as there are 6 different input values (such as wind velocity or humidity) that the model can use to make predictions.
	
	With Python, the LSTM model can be defined and compiled as follows:
	
	\begin{lstlisting}[basicstyle=\scriptsize,]
	multi_step_model = Sequential()
	multi_step_model.add(LSTM(32, return_sequences=True,
			input_shape=x_train_multi.shape[-2:]))
	multi_step_model.add(LSTM(16, activation='relu'))
	multi_step_model.add(Dense(n_output))
	
	multi_step_model.compile(optimizer='adam', loss='mse')
	\end{lstlisting}
	
	For this model, an LSTM network with 32 neurons in the first layer and 16 neurons in a hidden layer will be created. The Dense output layer contains a variable as the number of neurons. "n\_output" is the number of values that the model needs to predict. If the model would predict the temperature for 5 hours into the future, "n\_output" would be set to 5. 
	Here, the "ReLU" activation function makes sure that the outputs of the hidden layer are always positive.  
	"Adam" is an optimization algorithm, which updates the network weights in the training data. The Mean Squared Error (MSE) will be used as loss function for the training of the model. The MSE calculates the average squared difference between the estimated and the actual values. The training of the model happens in "epochs", which can be defined as one full pass over the whole training set. Intervals can be defined for the epochs. In each epoch, a "batch" is used for training, with one batch being equivalent to one sample. The model is updated, once all of the batches are processed. Finally, the amount of historical data that should be used by the model to make future predictions can be defined. The default value will be 168, equivalent to 7 days of data.
	
	Next, the temperature predictions are going to be conducted with the compiled model.
	\bigskip
	
	\subsection{Forecasting a 7-day period}
	In the first attempt, a multi step time series prediction for 7 days equivalent to 168 hours or 168 prediction points will be conducted. All 6 measures temp, hum, airpr, solrad, windvel, and winddir will be used.
	
	For this prediction, the following values are used for the model variables:
	\bigskip
	
	\begin{tabular}{l}
	batch size = 256 \\
	evaluation interval = 200 \\
	epochs = 10 \\
	historical data = 168  \\
	future steps = 168 \\
	\end{tabular}
	\bigskip
	
	After running the model, the training and validation loss can be evaluated by plotting the MSE values for both training and validation, such as in figure 9.
	
	\begin{figure}[!htbp]
		\centerline{\includegraphics[width=\linewidth]{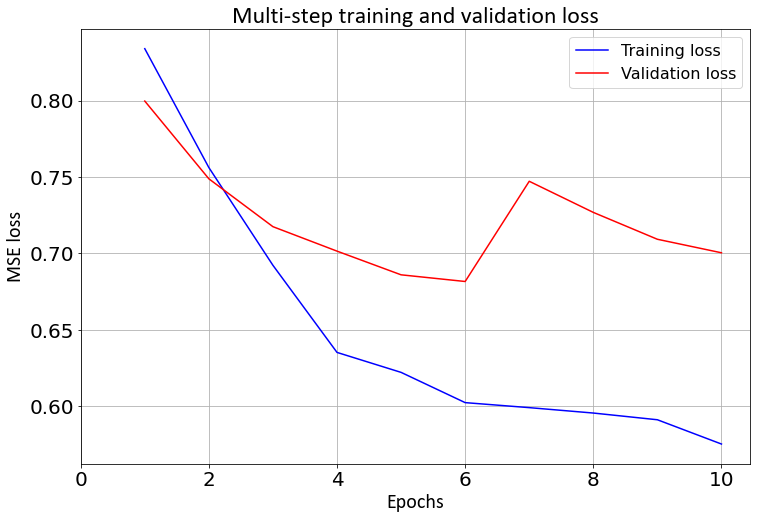}}
		\caption{MSE loss of the first run.}
		\label{fig}
	\end{figure}
	
	The diagram shows the number of epochs on the x-axis and the loss (or deviation) measured in MSE on the y-axis. The blue graph shows the trend of the training loss and the red graph the trend of the validation loss.
	
	A decrease of training and validation loss over time can be observed. However, after the second epoch until the end of the training, the validation loss is significantly higher than the training loss. When the validation loss during the training of a model stays significantly above the training loss, the model experiences overfitting. Overfitting refers to an ML model that learns the underlying data too well by modeling even the smallest details and noise in the training data, causing a high loss (or low accuracy) when testing on the validation data. In figure 9 a final MSE loss of 0.58 for training and 0.70 for validation of the model can be seen. 
	
	After compilation and training, the model can be deployed for a time series prediction.
	The following diagram shows a sample of a multi step time series prediction with 168 temperature points forecasted for 168 hours (or one week) into the future, using the prepared ML model from above:
	
	\begin{figure}[!htbp]
		\centerline{\includegraphics[width=\linewidth]{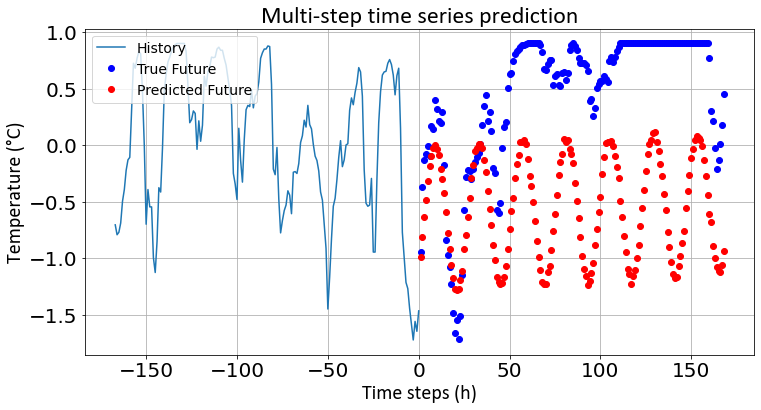}}
		\caption{7-day forecast after the first run.}
		\label{fig}
	\end{figure}

	The diagram shows the time steps in hours on the x-axis and the temperature measured in degrees Celsius on the y-axis. The graph progressing up to the time step 1 shows historical temperature data taken from the *.csv file. The red dots starting at time step 1 show the model’s prediction of the temperature for 168 hours following the historical temperature data. The blue dots show the real progression of temperature values following the historical temperature data. It can be observed that the model catches the progression of the true future values for the first 30 hours. The remaining predictions (especially after time step 50) do not match the true future values and are in some cases off by more than 2 degrees.
	
	The following subsection is aimed at decreasing the training and validation loss during training of the ML model and, if possible, increasing the overall accuracy of the predicted temperatures.
	\bigskip
	
	\subsection{Forecasting a 1-day period}
	
	For the second run, the LSTM model’s variables will be adjusted with the objective of increasing the accuracy of the model’s predictions. Here, a multi step time series prediction for 1 day equivalent to 24 hours or 24 prediction points will be conducted.
	
	For this prediction, the following values are used for the model variables:
	\bigskip
	
	\begin{tabular}{l}
	batch size = 512 \\
	evaluation interval = 100 \\
	epochs = 20 \\
	historical data = 168  \\
	future steps = 24 \\
	\end{tabular}
	\bigskip
	
	The historical data used by the model to make future predictions stays at 168 values. In an ideal scenario, the model’s training and validation loss will converge and keep decreasing with each epoch. That is why the number of epochs will be increased in this run by 10. The evaluation interval will be decreased to 100, minimizing the time for the run of each epoch and leading to a faster training result. The batch size will be increased to 512, providing the model a higher number of samples to do the training with.
	
	It has been mentioned in the previous run that a model that picks up noise tends to experience overfitting. One measure in the dataset that is particularly fluctuating and which could be picked up by the model as noise is winddir. The distribution of the values for winddir over the whole period of 9 years can be plotted, as shown in this figure:
	
	\begin{figure}[!htbp]
		\centerline{\includegraphics[width=\linewidth]{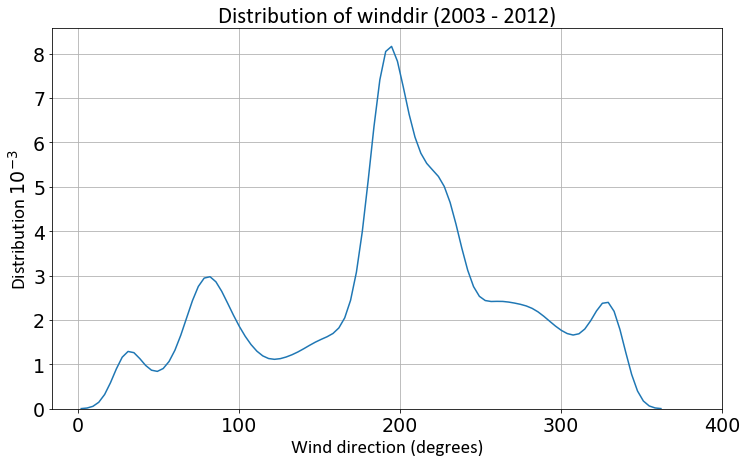}}
		\caption{Distribution of winddir over the period of 9 years.}
		\label{fig}
	\end{figure}
	
	It is apparent that the measure winddir is not distributed normally, which can be an obstacle for correlation analysis of the ML model.
	
	For this run, the measure winddir will be disregarded altogether due to the measure's volatility. The 5 remaining measures temp, hum, airpr, solrad, and windvel will be included. 
	Ideally, the tweak of the model’s variables, the prediction of a shorter time period, and the exclusion of the measure winddir will lead to a smoother training of the updated ML model and a better time series prediction.
	
	After running the model, the training and validation loss can once more be evaluated by plotting the MSE values for both training and validation, as seen in following figure:
	
	\begin{figure}[!htbp]
		\centerline{\includegraphics[width=\linewidth]{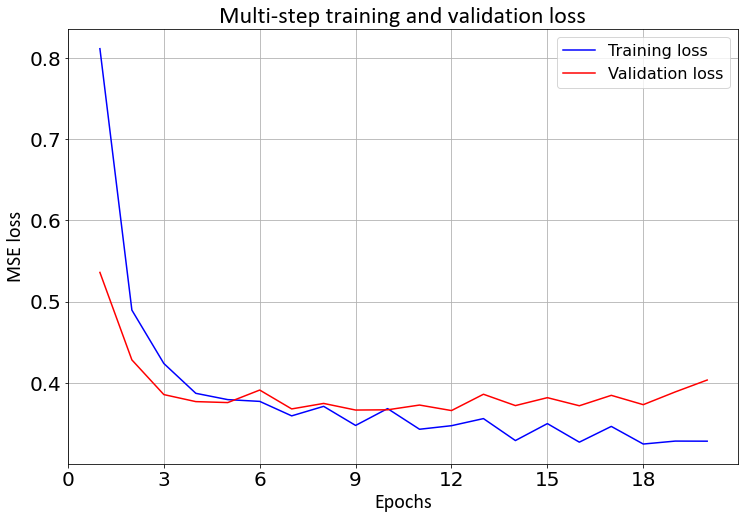}}
		\caption{MSE loss of the second run.}
		\label{fig}
	\end{figure}

	Figure 12 shows a drastic change in the training and validation loss compared to figure 9. In this instance, the final MSE loss stands at 0.33 for training and 0.40 for validation of the model, considerably lower than in the previous run. The two graphs converge better and the overfitting is not as pronounced as before. Nevertheless, an increase of validation loss towards the last epochs can be observed, which could indicate that the model is still overfitting the data.
	
	After compilation and training, the trained model can be deployed for a second time series prediction.
	
	The following diagram shows a sample of a multi step time series prediction with 24 temperature points forecasted for 24 hours (or one day) into the future:
	
	\begin{figure}[!htbp]
		\centerline{\includegraphics[width=\linewidth]{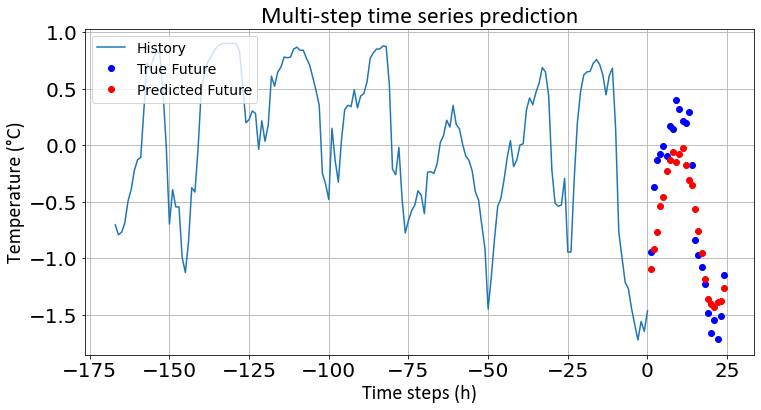}}
		\caption{1-day forecast after the second run.}
		\label{fig}
	\end{figure}

	In figure 13 it can be observed that the model catches the progression of all 24 future values fairly accurately. The fundamental shape formed by the predicted temperature points resembles the one of the true temperature points. The highest deviation seems to be around half a degree. In the updated model, the predicted temperature points are approaching the true future temperature points better than in the first run.
	
	In the following run, the previously improved model will be further optimized to obtain an even higher accuracy for the forecasted temperature.
	\bigskip
	
	\subsection{Forecasting a 12-hour period}
	
	For the last run, some of the LSTM model’s variables will be adjusted once more. Now, a multi step time series prediction for 12 hours equivalent to 12 prediction points will be conducted.
	
	For this prediction, the following values are used for the model variables:
	\bigskip
	
	\begin{tabular}{l}
	batch size = 512 \\
	evaluation interval = 150 \\
	epochs = 30 \\
	historical data = 48  \\
	future steps = 12 \\
	\end{tabular}
	\bigskip
	
	The historical data used by the model to make future predictions changes to 48 values, which will still be sufficient for a prediction but more importantly will accelerate the training of the model. Both the number of epochs and the evaluation interval increase, epochs to 30 and evaluation interval to 150. Again, the idea is to decrease the loss over time with a higher number of epochs. The batch size stays at 512.
	
	To tackle the issue with overfitting, that the last two versions of the LSTM model experienced, an "l2 regularizer" will be included in this updated LSTM network.
	
	Regularizers add penalties on different parameters of a model, which will decrease the freedom the model has when training with data. The generalization of models will be improved by suppressing the model’s ability to learn the noise of training data.
	
	The l2 regularizers can be added to the input layer of the LSTM model:
	
	\begin{lstlisting}[basicstyle=\scriptsize,]
	multi_step_model = tf.keras.models.Sequential()
	multi_step_model.add(tf.keras.layers.LSTM(32, 
			kernel_regularizer=l2(0.005), 
			recurrent_regularizer=l2(0.005), 
			bias_regularizer=l2(0.005), 
			return_sequences=True, 
			input_shape=x_train_multi.shape[-2:]))
	
	multi_step_model.add(tf.keras.layers.LSTM(16, 
			activation='relu'))
	multi_step_model.add(tf.keras.layers.Dense(12))
	
	multi_step_model.compile(optimizer='adam', loss='mse')
	\end{lstlisting}
	
	The regularizers are split into the kernel, the recurrent, and the bias regularizer, allowing for individual tuning of parameters. A higher regularizer value will lead to a higher model suppression. In this model, a value of 0.005 will be used.
	
	After running the updated model for a third time, the training and validation loss seen in figure 14 is output.
	
	\begin{figure}[!htbp]
		\centerline{\includegraphics[width=\linewidth]{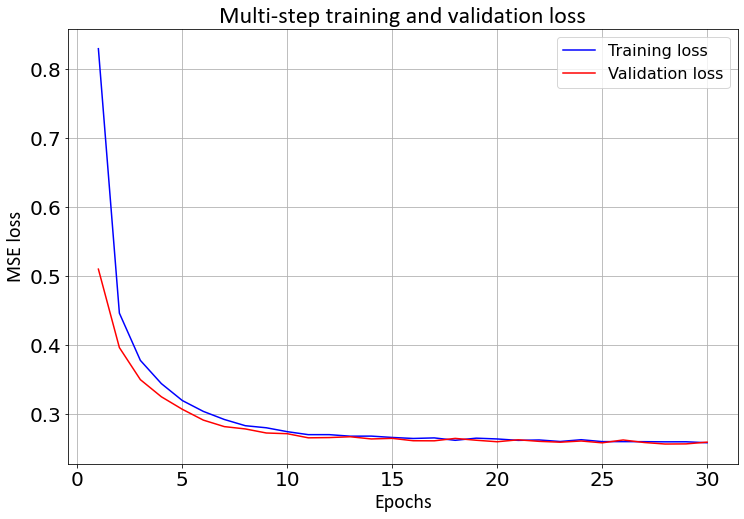}}
		\caption{MSE loss of the third run.}
		\label{fig}
	\end{figure}
	
	Figure 14 shows another improvement in training and validation loss over the previous training run. In this instance, the final MSE loss is 0.26 for training and 0.26 for validation of the model, again showing a decrease compared to the previous run. Particularly the validation loss has experienced a significant improvement, showing a decrease of 0.14 compared to the final MSE validation loss shown in figure 12.
	The two graphs‘ convergence is ideal and the overfitting from the previous two runs has been eliminated.
	
	The trained model can now be deployed for the last time series prediction.
	Figure 15 shows a sample of a multi step time series prediction with 12 temperature points forecasted for 12 hours into the future.

	\begin{figure}[!htbp]
		\centerline{\includegraphics[width=\linewidth]{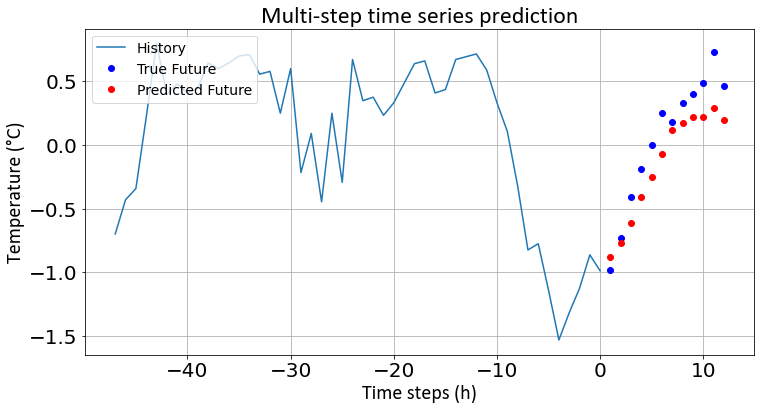}}
		\caption{12-hour forecast after the third run.}
		\label{fig}
	\end{figure}
	
	As seen in the diagram, the model catches the progression of the 12 future values with a high accuracy. The fundamental shape formed by the predicted temperature points resembles the one of the true temperature points. The highest deviation is less than 0.5 degrees. The updated model seems to show similar results as the model from the second run with a less significant improvement than in the previous update. 
	\bigskip
	
	\section{Results}

	For measuring the accuracy of the three LSTM model versions, the resulting predictions of the models can be compared with different measurements. Here, the root mean squared error (RMSE), mean average error (MAE), and maximum error value (ME) will be used as accuracy measures. 
	
	For the RMSE the quadratic mean of the differences between predicted values and true future values are calculated, before extracting the result in a root. The MAE is the average of all deviations between predicted and true values of a data set, and finally the ME shows the value with the biggest deviation between the predicted and true values in the set. 
	
	The following table shows the accuracy for each model measured in RMSE, MAE, and ME in degrees Celsius:
	\bigskip
	
	\begin{tabular}{l|l|l|l}
		
		Model & RMSE & MAE & ME \\\hline
		7-day forecast & 1.18 & 1.04 & 2.13 \\\hline
		1-day forecast & 0.34 & 0.29 & 0.63 \\\hline
		12-hour forecast & 0.24 & 0.21 & 0.44 \\
	\end{tabular}
	\bigskip
	
	For the 7-day forecast model, a value of around 1 for the measures RMSE and MAE can be observed. The ME is more than 2 degrees Celsius for this model. 
	After the tuning of parameters, the updated 1-day forecast model shows a drastic improvement in all 3 measures compared to the 7-day forecast model. The RMSE and MAE are both around 0.3, while the value with the highest deviation is 0.63 degrees Celsius in this case. 
	After the last update of the LSTM model, another slight improvement can be observed. The values for RMSE and MAE are around 0.1 units lower than in the 1-day forecast model. The ME is 0.44 degrees Celsius for the last model.
	
	This result shows that the accuracy of an LSTM model for a time series prediction can be improved by tuning the right parameters in the framework. By suppressing the noise, increasing the epochs, and decreasing the amount of values that need to be predicted by the model, a clear improvement in accuracy can be observed. This suggests that the model could be further tuned to decrease the errors during training and hence increase the accuracy of the predictions even more. Lastly, by using a shorter time period for the temperature prediction, a recursive multi step time series prediction model could be used to obtain more accurate results for predicting values further into the future.
	
	In this model, predicting the weather for multiple days appears to have its drawbacks, as the accuracy of the predicted values is low and the progression of the true future values is not captured correctly. Therefore, it is recommended to limit the number of prediction points to obtain a higher accuracy.
	\bigskip
	
	\section{Limitations}
	Contrary to the positive results shown in the example above, there are some fundamental limitations to time series predictions and more generally the application of neural networks that cannot be avoided by any optimization of RNNs. These limitations are going to be discussed in the next subsections. 
	\bigskip
	
	\subsection{Insufficient Quantity of Training Data}
	Compared to the neural network of a human brain, ANNs are not very developed yet. Using the example of image recognition, ANNs require thousands, if not millions of training examples to make good predictions. Humans, on the other hand can recognize and learn new images more efficiently [2]. 
	If the training data is insufficient, the ML algorithms might misinterpret test data, resulting in a higher error rate. Especially when the low amount of data creates an excessively simple model, ML frameworks tend to suffer from underfitting. Underfitting describes the inability of ML algorithms to capture underlying trends in a model. In summary, an insufficient quantity of training data usually leads to poor predictions of new data sets [17].	
	\bigskip
	
	\subsection{Nonrepresentative Training Data}
	For this subsection, a diagram from a nonrepresentative training will be shown.
	
	The diagram below plots a set of countries with the gross domestic product per capita in US dollars (GDP per capita (USD)) on the x-axis and the life satisfaction for each country on the y-axis. The blue, dotted line is the linear correlation line, disregarding the data for the countries Brazil, Mexico, Chile, Czech Republic, Norway, Switzerland, and Luxembourg. If the data is added for these 7 countries, the correlation line drastically changes, with a much flatter line compared to the dotted one. 
	
	\begin{figure}[!htbp]
		\centerline{\includegraphics[width=\linewidth]{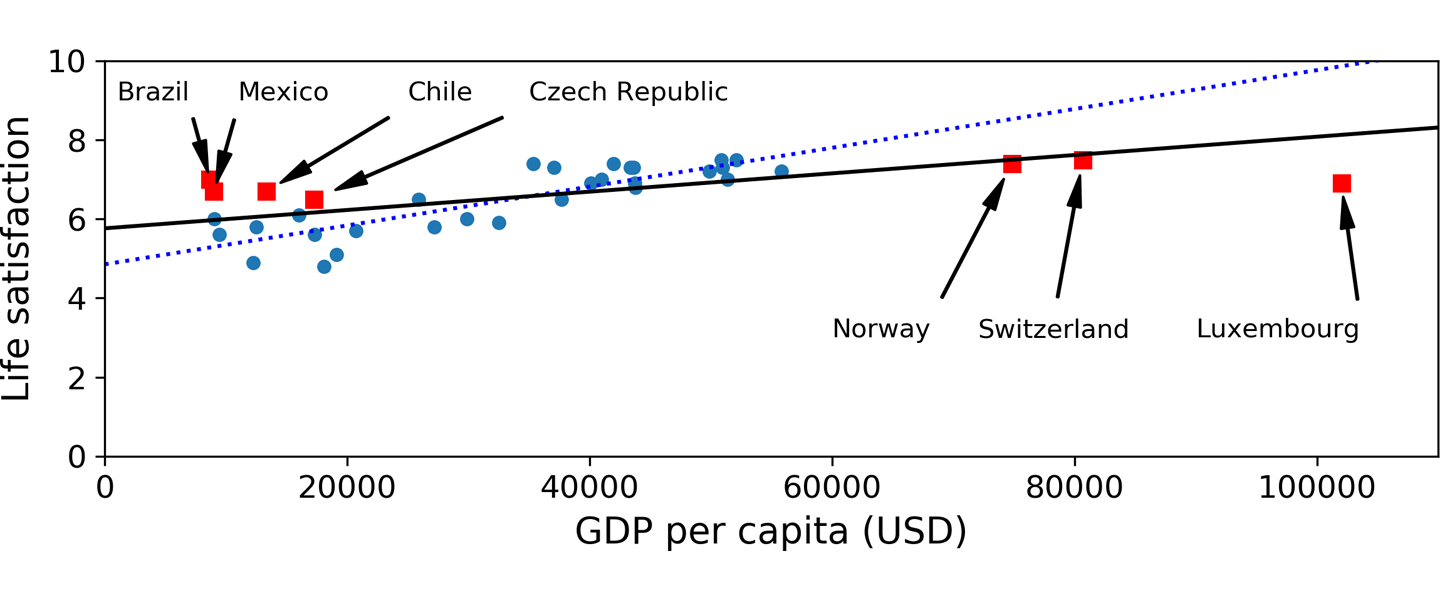}}
		\caption{A training sample showing the life satisfaction for different countries [2].}
		\label{fig}
	\end{figure}

	It is crucial for the data in a model to be representative, in order for ML algorithms to generalize well. This is important, so new cases that need to be predicted, can be generalized too [2].
	\bigskip
	
	\subsection{Poor-Quality Data}
	Maybe the most obvious limitation to ML is when the underlying model is not processed properly and its training data is full of errors and noise. Just like in the example of weather data prediction, it is important to prepare and clean the data before feeding it to the ML framework [2].
	The weather data that was used for the weather forecast contains the value wind direction, which ranges from 0 to 360. Because of the volatile nature of wind, the sequential values for wind direction show a very high fluctuation. If the fluctuation is too big, the ML framework could pick this information up as noise, which distorts the correlation between wind direction and other values. Disregarding or simplifying the values for such parameters could increase the quality of a model's result.

	\section{Conclusion}
	Precise weather forecasting is a complex study, which is based on weather observations, data processing, and correlation analysis. Generating accurate weather information in a sufficient amount for feeding ML frameworks requires a vast set of tools, models, and expertise. In this day and age, algorithms running on powerful machines output the information for weather predictions [18]. 
	Regardless of computing power, weather prediction involves limitations, which cannot be avoided with any finetuning of algorithms or precise modeling of RNNs. On average, weather forecasts come with an accuracy of 80\% for a seven-day forecast, and an accuracy of less than 50\% for a forecast of 10 or more days [19]. 
	
	The same limitation could be observed with the LSTM model, which was used to make the weather prediction for this paper. It has been proven [10] that LSTM models tend to get worse with time. Further, the model does not perform very well in non-stationary, high noise scenarios. These scenarios can generally be avoided by preparing the underlying data and tuning the parameters of the model (e.g. the layers in the LSTM network or the learning rate) to increase the accuracy and quality of the prediction. The volatile nature of weather makes for an additional obstacle, which often cannot be tackled easily with the LSTM model.
	
	In the above tests, an improvement of temperature forecasts with the tuning of the LSTM model’s parameters could be observed. Additional data such as information from other weather stations or satellite pictures showing the course of rain clouds and storms could be helpful for making a precise temperature forecast for a longer time period, such as for 7 days. Without these additional tools, the limitations of the LSTM model pose a challenge for precise and effective weather forecasts. To tackle these limitations, models can be finetuned and tested with various parameter values. 
	In particular the use of more complex frameworks, such as the recursive multi step time series model should be taken into consideration. Because LSTM models are showing a higher precision for shorter time series, including each predicted value in the following runs of a recursive model could be highly profitable.
	
	All in all, the TensorFlow framework with the Keras extension provides a very powerful and easy to use tool to make time series predictions. Particularly creating an artificial neural network with Keras using the LSTM model for a short time period shows very convincing results, when testing the trained framework. The above result suggests that the weather predictions can be optimized even further with the right tuning of parameters.

\end{document}